%% file: acl2015.tex
\documentclass[11pt]{article}
\usepackage{acl2015}
\usepackage{times}
\usepackage{url}
\usepackage{latexsym}
\usepackage{amssymb}
\usepackage{amsmath}
\usepackage{mathtools}
\usepackage{graphicx}
\usepackage{enumitem}
\usepackage{multirow}

\title{Statistical Machine Translation Features with Multitask Tensor Networks }
\author{Hendra Setiawan, Zhongqiang Huang, Jacob Devlin$^\dag$\thanks{* Research conducted when the author was at BBN.}\,\,,\,Thomas Lamar, \\ \textbf{Rabih Zbib, Richard Schwartz and John Makhoul}\\
Raytheon BBN Technologies, 10 Moulton St, Cambridge, MA 02138, USA\\
$^\dag$Microsoft Research, One Microsoft Way, Redmond, WA 98052, USA\\
 \texttt{\{hsetiawa,zhuang,tlamar,rzbib,schwartz,makhoul\}@bbn.com}\\ \texttt{jdevlin@microsoft.com}
}
\date{}

\begin{document}

\maketitle

\input{abstract}
\input{introduction}

\input{models}

\input{tensor-networks}

\input{multitask_learning}
\input{experiments}
\input{related_work}
\input{discussion}

\section*{Acknowledgement}
This work was supported by DARPA/I2O Contract No. HR0011-12-C-0014 under the BOLT Program. The views, opinions, and/or findings contained in this article are those of the author and should not be interpreted as representing the official views or policies, either expressed or implied, of the Defense Advanced Research Projects Agency or the Department of Defense.

\bibliographystyle{acl}
\bibliography{acl2015}

\end{document}

%% file: abstract.tex
\begin{abstract}

We present a three-pronged approach to improving Statistical Machine Translation (SMT), building on recent success in the application of neural networks to SMT. 
First, we propose new \emph{features} based on neural networks to model various non-local translation phenomena. 
Second, we augment the \emph{architecture} of the neural network with tensor layers that capture important higher-order interaction among the network units. 
Third, we apply multitask \emph{learning} to estimate the neural network parameters jointly. 
Each of our proposed methods results in significant improvements that are complementary.
The overall improvement is +2.7 and +1.8 BLEU points for Arabic-English and Chinese-English translation over a state-of-the-art system that already includes neural network features.

\end{abstract}

%% file: introduction.tex
\section{Introduction}

Recent advances in applying Neural Networks to Statistical Machine Translation (SMT) have generally taken one of two approaches. They either develop neural network-based features that are used to score hypotheses generated from traditional translation grammars \cite{sundermeyer-EtAl:2014:EMNLP2014,devlin-EtAl:2014:P14-1,auli2013,le2012,schwenk2012}, or they implement the whole translation process as a single neural network \cite{bahdanauChoBengio,Ilya_NIPS2014_5346}. 
The latter approach, sometimes referred to as \emph{Neural Machine Translation}, attempts to overhaul SMT, while the former capitalizes on the strength of the current SMT paradigm and leverages the modeling power of neural networks to improve the scoring of hypotheses generated by phrase-based or hierarchical translation rules. This paper adopts the former approach, as \emph{n}-best scores from state-of-the-art SMT systems often suggest that these systems can still be significantly improved with better features.

We build on \cite{devlin-EtAl:2014:P14-1} who proposed a simple yet powerful feedforward neural network model that estimates the translation probability conditioned on the target history and a large window of source word context. We take advantage of neural networks' ability to handle sparsity, and to infer useful abstract representations automatically. At the same time, we address the challenge of learning the large set of neural network parameters. In particular,

\begin{itemize}[noitemsep,nolistsep]
\item We develop new {\em Neural Network Features} to model non-local translation phenomena related to word reordering. Large fully-lexicalized contexts are used to model these phenomena effectively, making the use of neural networks essential. All of the features are useful individually, and their combination results in significant improvements (Section \ref{s_many_features}).
\item We use a {\em Tensor Neural Network Architecture} \cite{conf/interspeech/YuDS12} to automatically learn complex pairwise interactions between the network nodes. The introduction of the tensor hidden layer results in more powerful features with lower model perplexity and significantly improved MT performance for all of neural network features (Section \ref{s_dtn}). 
\item We apply {\em Multitask Learning} (MTL) \cite{c-ml-97} to jointly train related neural network features by sharing parameters. This allows parameters learned for one feature to benefit the learning of the other features. 
This results in better trained models and achieves additional MT improvements (Section \ref{s_mtl}).
\end{itemize}

We apply the resulting \emph{Multitask Tensor Networks} to the new features and to existing ones, obtaining strong experimental results over the strongest previous results of \cite{devlin-EtAl:2014:P14-1}. We obtain improvements of +2.5 BLEU points for Arabic-English and +1.8 BLEU points for Chinese-English on the DARPA BOLT Web Forum condition. We also obtain improvements of +2.7 BLEU point for Arabic-English and +1.9 BLEU points for Chinese-English on the NIST Open12 test sets over the best previously published results in \cite{devlin-EtAl:2014:P14-1}. Both the tensor architecture and multitask learning are general techniques that are likely to benefit other neural network features.

%% file: models.tex
\section{New Non-Local SMT Features}
\label{s_many_features}

Existing SMT features typically focus on local information in the source sentence, in the target hypothesis, or both. For example, the \emph{n}-gram language model (LM) predicts the next target word by using previously generated target words as context (local on target), while the lexical translation model (LTM) predicts the translation of a source word by taking into account surrounding source words as context (local on source). 

In this work, we focus on non-local translation phenomena that result from non-monotone reordering, where \emph{local} context becomes \emph{non-local} on the other side. We propose a new set of powerful MT features that are motivated by this simple idea.
To facilitate the discussion, we categorize the features into {\em hypothesis-enumerating} features that estimates a probability for each generated target word (e.g., $n$-gram language model), and {\em source-enumerating} features that estimates a probability for each source word (e.g., lexical translation). 

More concretely, we introduce a) \emph{Joint Model with Offset Source Context} (JMO), a hypothesis enumerating feature that predicts the next target word the source context affiliated to the previous target words; and b) \emph{Translation Context Model} (TCM), a source-enumerating feature that predicts the context of the translation of a source word rather than the translation itself. 
These two models extend pre-existing features: the Joint (language and translation) Model (JM) of \cite{devlin-EtAl:2014:P14-1} and the LTM respectively respectively. 
We use a large lexicalized context for there features, making the choice of implementing them as neural networks essential.
We also present neural-network implementations of pre-existing source-enumerating features: lexical translation, orientation and fertility models. We obtain additional gains from using tensor networks and multitask learning in the modeling and training of all the features.

\subsection{Hypothesis-Enumerating Features}

As mentioned, hypothesis-enumerating features score each word in the hypothesis, typically by conditioning it on a context of $n$-1 previous target words as in the $n$-gram language model.
One recent such model, the joint model of \newcite{devlin-EtAl:2014:P14-1} achieves large improvements to the state-of-the-art SMT by using a large context window of 11 source words and 3 target words.
The Joint Model with Offset Source Context (JMO) is an extension of the JM that uses the source words affiliated with the $n$-gram target history as context. 
The source contexts of JM and JMO overlap highly when the translation is monotone, but are complementary when the translation requires word reordering. 

\subsubsection{Joint Model with Offset Source Context}
\label{ss_jmo}

Formally, JMO estimates the probability of the target hypothesis $E$ conditioned on the source sentence $F$ and a target-to-source \textit{affiliation} $\boldsymbol{A}$:
\begin{equation}
P(E|F,\boldsymbol{A}) \approx \prod_{i=1}^{|E|} P(e_i | e_{i-1}^{i-n+1}, \mathcal{C}_{a_{i-k}}=f_{a_{i-k}-m}^{a_{i-k}+m}) \nonumber
\end{equation}
\noindent where $e_i$ is the word being predicted; $e_{i-1}^{i-n+1}$ is the string of $n-1$ previously generated words; $\mathcal{C}_{a_{i-k}}$ to the source context of $m$ source words around $f_{a_{i-k}}$, the source word affiliated with $e_{i-k}$. We refer to $k$ as the offset parameter. We use the definition of {\em word affiliation} introduced in \newcite{devlin-EtAl:2014:P14-1}. When no source context is used, the model is equivalent to an $n$-gram language model, while an offset parameter of $k=0$ reduces the model to the JM of \newcite{devlin-EtAl:2014:P14-1}.

When $k>0$, the JMO captures non-local context in the prediction of the next target word.
More specifically, $e_{i-k}$ and $e_i$, which are local on the target side, are affiliated to $f_{a_{i-k}}$ and $f_{a_{i}}$ which may be distant from each other on the source side due to non-monotone translation, even for $k=1$. 
The offset model captures reordering constraints by encouraging the predicted target word $e_i$ to fit well with the previous affiliated source word $f_{a_{i-k}}$ and its surrounding words. 
We implement a separate feature for each value of $k$, and later train them jointly via multitask learning. 
As our experiments in Section~\ref{ss_novel_features_effects} confirm, the history-affiliated source context results in stronger SMT improvement than just increasing the number of surrounding words in JM. 

\begin{figure}[thb]
\begin{center}
\includegraphics[width=\columnwidth]{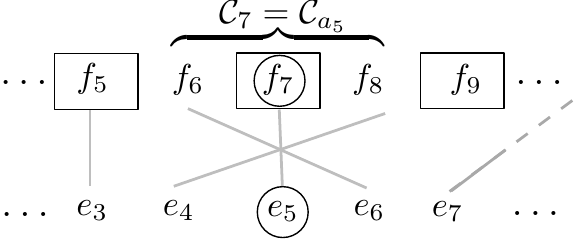}
\end{center}
\caption{Example to illustrate features. $f_5^9$ is the source segment, $e_3^7$ is the corresponding translation and lines refer to the alignment. We show hypothesis-enumerating features that look at $f_7$ and source-enumerating features that look at $e_5$. We surround the source words affiliated with $e_5$ and its $n$-gram history with a bracket, and surround the source words affiliated with the history of $e_5$ with squares. }\label{f_example}
\end{figure}

Fig.~\ref{f_example} illustrates the difference between JMO and JM. Assuming $n=3$ and $m=1$, then JM estimates $P(e_5 | e_4, e_3, \mathcal{C}_{a_5}=\{f_6, f_7, f_8\})$. On the other hand, for $k=1$ , JMO$_{k=1}$ estimates $P(e_5 | e_4, e_3, \mathcal{C}_{a_4}=\{f_8, f_9, f_{10}\})$.

\subsection{Source-Enumerating Features}
\label{s_translation_orientation}

Source-Enumerating Features iterate over words in the source sentence, including unaligned words, and assign it a score depending on what aspect of translation they are modeling.
A source-enumerating feature can be formulated as follows:
\begin{equation}
P(E|F,\boldsymbol{A}) \approx \prod_{j=1}^{|F|} P(Y_j | \mathcal{C}_{j}=f_{j-m}^{j+m}) \nonumber
\end{equation}
where $\mathcal{C}_{a_j}$ is the source context (similar to the hypothesis-enumerating features above) and $Y_j$ is the label being predicted by the feature. 
We first describe pre-existing source-enumerating features: the lexical translation model, the orientation model and the fertility model, and then discuss a new feature: Translation Context Model (TCM), which is an extension of the lexical translation model.


\subsubsection{Pre-existing Features}

\emph{Lexical Translation model} (LTM) estimates the probability of translating a source word $f_j$ to a target word $l(f_j)=e_{b_j}$ given a source context $\mathcal{C}_j$, $b_j\in \boldsymbol{B}$ is the source-to-target word affiliation as defined in \cite{devlin-EtAl:2014:P14-1}. When $f_j$ is translated to more than one word, we arbitrarily keep the left-most one. The target word vocabulary $V$ is extended with a $NULL$ token to accommodate unaligned source words.

\emph{Orientation model} (ORI) describes the probability of orientation of the translation of phrases surrounding a source word $f_j$ relative to its own translation.
We follow \cite{setiawan-EtAl:2013:ACL2013} in modeling the orientation of the left and right phrases of $f_j$ with maximal orientation span (the longest neighboring phrase consistent with alignment), which we denote by $L_j$ and $R_j$ respectively. Thus, $o(f_j)=\langle o_{L_j}(f_j), o_{R_j}(f_j) \rangle$, where $o_{L_j}$ and $o_{R_j}$ refer to the orientation of $L_j$ and $R_j$ respectively. For unaligned $f_j$, we set $o(f_j) = o_{L_j}(R_j)$, the orientation of $R_j$ with respect to $L_j$. 


\emph{Fertility model} (FM) models the probability that a source word $f_j$ generates $\phi(f_j)$ words in the hypothesis. Our implemented model only distinguishes between aligned and unaligned source words (i.e., $\phi(f_j) \in \{0,1\}$). The generalization of the model to account for multiple values of $\phi(f_i)$ is straightforward.

\subsubsection{Translation Context Model}

As with JMO in Section \ref{ss_jmo}, we aim to capture translation phenomena that appear local on the target hypothesis but non-local on the source side.  Here, we do so by extending the LTM feature to predict not only the translated word $e_{b_j}$, but also its surrounding context. 
Formally, we model $P(l(f_j)|\mathcal{C}_j)$, where $l(f_j)=e_{b_j-d}, \cdots, e_{b_j}, \cdots e_{b_j+d}$ is the hypothesis word window around $e_{b_j}$.
In practice, we decompose TCM further into $\prod\limits^{+d}_{d' = -d} P(e_{b_j+d'}|\mathcal{C}_j)$ and implemented each as a separate neural network-based feature. Note that TCM is equivalent to the LTM when $d=0$. Because of word reordering, a given hypothesis word in $l(f_j)$ might not be affiliated with $f_j$ or even to the words in $\mathcal{C}_j$. TCM can model non-local information in this way.

\subsubsection{Combined Model}

Since the feature label is undefined for unaligned source words, we make the model hierarchical, based on whether the source word is aligned or not, and thus arrive at the following formulation:
\begin{equation}
\begin{array}{l}
\mkern-28mu P(l(f_j)) \cdot P(ori(f_j)) \cdot P(\phi(f_j))= \\ [6pt]
\,\left\{\begin{array}{l}
            \!\!\!P(\phi_p(f_j)=0) \cdot P(o_{L_j}(R_j)) \\ [8pt]
            \!\!\!P(\phi_p(f_j) \ge 1) \cdot \prod\limits^{+d}_{d' = -d} P(e_{b_j+d'}) \\ 
            \,\,\cdot P(o_{L_j}(f_j),o_{R_j}(f_j))
       \end{array}
       \right. \nonumber
\end{array}
\end{equation}
\noindent We dropped the common context ($\mathcal{C}_j$) for readability. 



We reuse Fig.~\ref{f_example} to illustrate the source-enumerating features. Assuming $d=1$, the scores associated with $f_7$ are $P(\phi(f_7) \ge 1|\mathcal{C}_7)$ for the FM; $P(e_4|\mathcal{C}_7) \cdot P(e_5|\mathcal{C}_7) \cdot P(e_6)|\mathcal{C}_7)$ for the TCM; 
and $P(o(f_7) = \langle o_{L_7}(f_7) = RA, o_{R_7}(f_7)=RA\rangle)$ for the ORI($RA$ refers to {\em Reverse Adjacent}). $L_7$ and $R_7$ (i.e. $f_{6}$ and $f_8^9$ respectively), the longest neighboring phrase of $f_7$, are translated in reverse order and adjacent to $e_5$.

%% file: tensor-networks.tex
\section{Tensor Neural Networks}
\label{s_dtn}

The second part of this work improves SMT by improving the neural network architecture. 
Neural Networks derive their strength from their ability to learn a high-level representation of the input automatically from data. This high-level representation is typically constructed layer by layer through a weighted sum linear operation and a non-linear activation function. 
With sufficient training data, neural networks often achieve state-of-the-art performance on many tasks. This stands in sharp contrast to other algorithms that require tedious manual feature engineering. 
For the features presented in this paper, the context words are fed to the network network with minimal engineering. 

We further strengthen the network's ability to learn rich interactions between its units by introducing tensors in the hidden layers. 
The multiplicative property of the tensor bares a close resemblance to collocation of context words which are useful in many natural language processing tasks.


In conventional feedforward neural networks, the output of hidden layer $l$ is produced by multiplying the output vector from the previous layer with a weight matrix $(W_{l})$ and then applying the activation function $\sigma$ to the product. 
Tensor Neural Networks generalize this formulation by using a tensor $U_{l}$ of order 3 for the weights. The output of node $k$ in layer $l$ is computed as follows:
\begin{equation}
h_{l}[k] = \sigma\left( h_{l-1} \cdot U_{l}[k] \cdot h_{l-1}^T\right) \nonumber
\end{equation}
\noindent where $U_{l}[k]$, the $k$-th slice of $U_l$, is a square matrix.

In our implementation, we follow \cite{conf/interspeech/YuDS12,Hutchinson:2013:TDS:2498740.2498878} and use a low-rank approximation of $U_{l}[k] = Q_{l}[k] \cdot R_{l}[k]^{T}$, where $Q_{l}[k], R_{l}[k] \in \mathbb{R}^{n \times r}$. The output of node $k$ becomes: 
\begin{equation}
h_{l}[k] = \sigma\left( h_{l-1} \cdot Q_{l}[k] \cdot R_{l}[k]^{T} \cdot h_{l-1}^{T}\right) \nonumber
\end{equation}
\noindent 

In our experiments, we choose $r=1$, and also apply the non-linear activation function $\sigma$ distributively.
We arrive at the following three equations for computing the hidden layer outputs $(0<l<L)$:
\begin{eqnarray}
v_{l} & = & \sigma\left( h_{l-1} \cdot Q_{l}\right)  \nonumber \\
v^\prime_{l} & = & \sigma\left( h_{l-1} \cdot R_{l}\right) \nonumber\\
h_{l} & = & v_{l} \otimes v^\prime_{l} \nonumber
\end{eqnarray}
\noindent where $h_{l-1}$ is double-projected to $v_l$ and $v^\prime_{l}$, and the two projections are merged using the Hadamard element-wise product operator $\otimes$.

\begin{figure*}
\begin{center}
\label{f_network}
\includegraphics{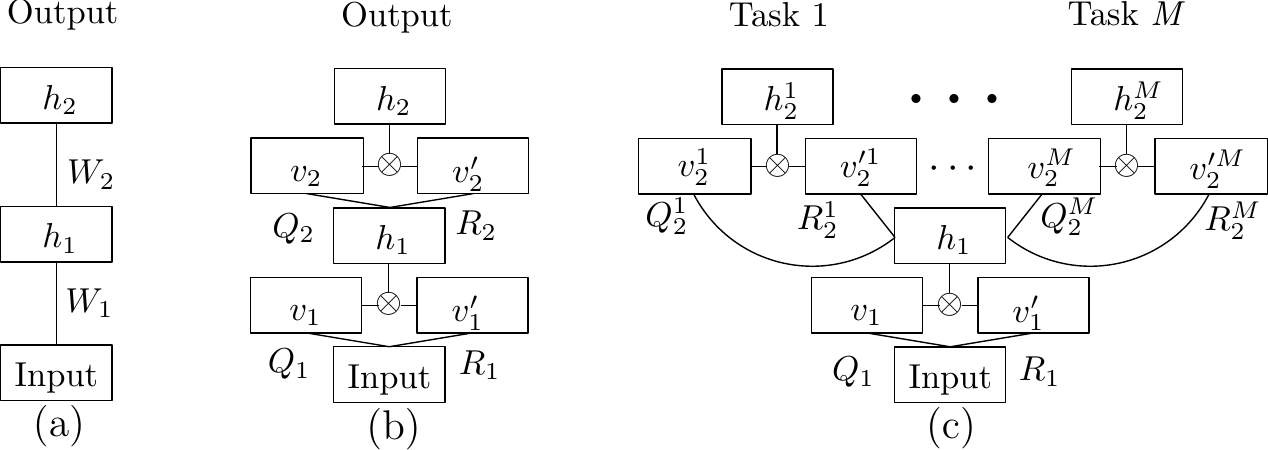}
\end{center}
\caption{The network architecture for (a) a conventional feedforward neural network, (b) tensor hidden layers, and (c) multitask learning with $M$ features that share the embedding and first hidden layers ($t=1$).}
\end{figure*}

This formulation allows us to use the same infrastructure of the conventional neural networks by projecting the previous layer to two different spaces of the same dimensions, then multiplying them element-wise. 
The only component that is different from conventional feedforward neural networks is the multiplicative function, which is trivially differentiable with respect to the learnable parameters. 
Figure~\ref{f_network}(b) illustrates the tensor architecture for two hidden layers.

The tensor network can learn collocation features more easily. For example, it can learn a collocation feature that is activated only if $h_{l-1}[i]$ collocates with $h_{l-1}[j]$ by setting $U_{l}[k][i][j]$ to some positive number. This results in SMT improvements as we describe in Section~\ref{s_experiments}.



%% file: multitask_learning.tex
\section{Multitask Learning}
\label{s_mtl}

The third part of this paper addresses the challenge of effectively learning a large number of neural network parameters without overfitting.
The challenge is even larger for tensor network since they practically doubles the number of parameters. 
In this section, we propose to apply Multitask Learning (MTL) to partially address this issue. We implement MTL as parameter sharing among the networks. This effectively reduces the number of parameters, and more importantly, it takes advantage of parameters learned for one feature to better learn the parameters of the other features. Another way of looking at this is that MTL facilitates regularization through learning the other tasks.


MTL is suitable for SMT features as they model different but closely related aspects of the same translation process. 
MTL has long been used by the wider machine learning community \cite{c-ml-97} and more recently for natural language processing \cite{Collobert:2008,Collobert:2011}. 
The application of MTL to machine translation, however, has been much less restricted, which is rather surprising since SMT features arise from the same translation task and are naturally related. 


We apply MTL for the features described in Section \ref{s_many_features}. We design all the features to also share the same neural network architecture (in this case, the tensor architecture described in Section~\ref{s_dtn}) and the same input, thus resulting in two large neural networks: one for the hypothesis-enumerating features and another for the source-enumerating ones. 
This simplifies the implementation of MTL.
Using this setup, it is possible to vary the number of shared hidden layers $t$ from 0 (only sharing the embedding layer) to $L-1$ (sharing all the layers except the output). Note that in principle MTL is applicable to other set of networks that have different architecture or even different input set. With MTL, the training procedure is the same as that of standard neural networks. 

We use the back propagation algorithm, and use as the loss function the product of likelihood of each feature\footnote{In this and in the other parts of the paper, we add the normalization regularization term described in \cite{devlin-EtAl:2014:P14-1} to the loss function to avoid computing the normalization constant at model query/decoding time.}:

\begin{equation}
Loss = \sum_i \sum_j^M \log \left( P\left(Y_j(X_i)\right) \right) \nonumber
\end{equation}
\noindent where $X_i$ is the training sample and $Y_j$ is one of the $M$ models trained. We use the sum of log likelihoods since we assume that the features are independent.

Fig.~\ref{f_network}(c) illustrates MTL between $M$ models sharing the input embedding layer and the first hidden layer ($t=1$) compared to the separately-trained conventional feedforward neural network and tensor neural network.

%% file: experiments.tex
\section{Experiments}
\label{s_experiments}

We demonstrate the impact of our work with extensive MT experiments on Arabic-English  and Chinese-English translation for the DARPA BOLT Web Forum and the NIST OpenMT12 conditions. 

\subsection{Baseline MT System}

We run our experiments using a state-of-the-art string-to-dependency hierarchical decoder \cite{shen2010}. The baseline we use includes a set of powerful features as follow:
\begin{itemize}[noitemsep,nolistsep]
\item Forward and backward rule probabilities
\item Contextual lexical smoothing \cite{devlin2009}
\item 5-gram Kneser-Ney LM
\item Dependency LM  \cite{shen2010}
\item Length distribution \cite{shen2010}
\item Trait features \cite{devlin2012}
\item Factored source syntax \cite{huang2013}
\item Discriminative sparse feature, totaling 50k features \cite{chiang2009}
\item Neural Network Joint Model (NNJM) and Neural Network Lexical Translation Model (NNLTM) \cite{devlin-EtAl:2014:P14-1}
\end{itemize}
As shown, our baseline system already includes neural network-based features. NNJM, NNLTM and use two hidden layers with 500 units and use embedding of size 200 for each input. 


We use the MADA-ARZ tokenizer \cite{habash2013} for Arabic word tokenization. For Chinese tokenization, we use a simple longest-match-first lexicon-based approach. We align the training data using GIZA++ \cite{och2003}. For tuning the weights of MT features including the new features, we use iterative $k$-best optimization with an ExpectedBLEU objective function \cite{rosti2010}, and decode the test sets after 5 tuning iteration. We report the lower-cased BLEU and TER scores.

\subsection{BOLT Discussion Forum}

The bulk of our experiments is on the BOLT Web Discussion Forum domain, which uses data collected by the LDC. The parallel training data consists of all of the high-quality NIST training corpora, plus an additional 3 million words of translated forum data. The tuning and test sets consist of roughly 5000 segments each, with 2 independent references for Arabic and 3 for Chinese.

\subsubsection{Effects of New Features}
\label{ss_novel_features_effects}

We first look at the effects of the proposed features compared to the baseline system. Table~\ref{table:models} summarizes the primary results of the Arabic-English and Chinese-English experiments for the BOLT condition. We show the experimental results related to hypothesis-enumerating features (HypEn) in rows $S_2$-$S_5$, those related to source-enumerating features (SrcEn) in rows $S_6$-$S_9$, and the combination of the two in row $S_{10}$.  For all the features, we set the source context length to $m=5$ (11-word window). For JM and JMO, we set the target context length to $n=4$. For the offset parameter $k$ of JMO, we use values 1 to 3. For TCM, we model one word around the translation ($d=1$). Larger values of $d$ did not result in further gains. The baseline is comparable to the best results of \cite{devlin-EtAl:2014:P14-1}. 

In rows $S_3$ to $S_5$, we incrementally add a model with different offset source context, from $k=1$ to $k=3$. For AR-EN, adding JMOs with different offset source context consistently yields positive effects in BLEU score, while in ZH-EN, it yields positive effects in TER score. Utilizing all offset source contexts ``+JMO$_{k\leq3}$'' (row $S_5$) yields around 0.9 BLEU point improvement in AR-EN and around 0.3 BLEU in ZH-EN compared to the baseline. The JMO consistently provides better improvement compared to a larger JM context (row $S_2$), validating our hypothesis that using offset source context captures important non-local context.

Rows $S_6$ to $S_9$ present the improvements that result from implementing pre-existing source-enumerating SMT features as neural networks, and highlight the contribution of our translation context model (TCM). 
This set of experiments is orthogonal to the HypEn experiments (rows $S_2$-$S_5$). Each pre-existing model has a modest positive cumulative effect on both BLEU and TER. We see this result as further confirming the current trend of casting existing SMT features as neural network since our baseline already contains such features. The next row present the results of adding the translation context model, with one word surrounding the translation ($d=1$). As shown, TCM yields a positive effect of around 0.5 BLEU and TER improvements in AR-EN and around 0.2 BLEU and TER improvements in ZH-EN. 

Separately, the set of source-enumerating features and the set of target-enumerating features produce around 1.1 to 1.2 points BLEU gain in AR-EN and 0.3 to 0.5 points BLEU gain in ZH-EN.
The combination of the two sets produces a complementary gain in addition to the gains of the individual models as Row ($S_{10}$) shows.
The combined gain improves to 1.5 BLEU points in AR-EN and 0.7 BLEU points in ZH-EN. 

\begin{table}[htb]
\begin{center}
\begin{tabular}{| l | r | r | r | r |}
\hline
  \multirow{2}{*}{System} & \multicolumn{2}{c|}{AR-EN} & \multicolumn{2}{c|}{ZH-EN} \\
  \cline{2-5}
                     & BL & TER & BL & TER \\
\hline
\hline
 $S_1$: Baseline           & 43.2 & 45.0 & 30.2 & 58.3\\
\hline
 $S_2$: $S_1$+JM$_{LC_8}$           & 43.5 & 45.0 & 30.2 & 58.5\\
\hline
\hline
 $S_3$: $S_1$+JMO$_{k=1}$            & 43.9 & 44.7 & 30.8 & 57.8\\
 $S_4$: $S_3$+JMO$_{k=2}$            & 43.9 & 44.7 & 30.7 & 57.8\\
 $S_5$: $S_4$+JMO$_{k=3}$            & 44.4 & 44.5 & 30.5 & 57.5\\
\hline
\hline
 $S_6$: $S_1$+LTM                    & 43.5 & 44.7 & 30.3 & 58.0\\
 $S_7$: $S_6$+ORI                    & 43.7 & 44.6 & 30.4 & 57.8\\
 $S_8$: $S_7$+FERT                   & 43.8 & 44.7 & 30.5 & 57.8\\
 $S_9$: $S_8$+TCM                    & 44.3 & 44.2 & 30.7 & 57.5\\
\hline\hline
 $S_{10}$: $S_9$+JMO$_{k\leq 3}$    & 44.7 & 44.1 & 30.9 & 57.3\\
\hline
\end{tabular}
\end{center}
\caption{MT results of various model combination in BLEU and in TER.\label{table:models}}
\end{table}


\subsubsection{Effects of Tensor Network and Multitask Learning}
\label{ss_dtn_mtl_effects}

We first analyze the impact of tensor architecture and MTL intrinsically by reporting the models' average log-likelihood on the validation sets (a subset of the test set) in Table~\ref{t_intrinsic}. As mentioned, we group the models to HypEn (JM and JMO$_{k\leq3}$) and SrcEn (LTM, ORI,FERT and TCM) as we perform MTL on these two groups. Likelihood of these two groups in the previous subsection are in column ``NN'' (for Neural Network), which serves as a baseline.
The application of the tensor architecture improves their likelihood as shown in column ``Tensor'' for both languages and models.

\begin{table}[htb]
\begin{tabular}{| l | l | r | r | r | r |}
\hline
                    & \multirow{3}{*}{Feat.} & \multicolumn{2}{c|}{Independent} & \multicolumn{2}{c|}{MTL} \\
  \cline{3-6}
                    & & NN & Tensor & $t=0$ & $t=1$ \\
                    & &          &         & $L=2$  & $L=3$ \\
\hline
\parbox[t]{2mm}{\multirow{2}{*}{\rotatebox[origin=c]{90}{AR}}} & HypEn & -8.85 & -8.54 & -8.35 & - \\
& SrcEn & -8.47 & -8.32 & -8.10 & -8.09 \\
\hline
\parbox[t]{2mm}{\multirow{2}{*}{\rotatebox[origin=c]{90}{ZH}}} & HypEn & -11.48 & -11.06 & -10.87 & - \\
& SrcEn & -10.77 & -10.66 & -10.54 & -10.49 \\
\hline
\end{tabular}
\caption{Sum of the average log-likelihood of the models in HypEn and SrcEn. $t=0$ refers to MTL that shares only the embedding layer, while $t=1$ shares the first hidden layer as well. $L$ refers to the network's depth. Higher value is better.\label{t_intrinsic}}
\end{table}

The likelihoods of the MTL-related experiments are in columns with ``MTL'' header. We present two set of results. In the first set (column ``MTL,t=0,L=2''), we run MTL for features from column ``Tensor'' by sharing the embedding layer only ($t=0$). This allows us to isolate the impact of MTL in the presence of Tensors. Column ``MTL,t=1,l=3'' corresponds to the experiment that produces the best intrinsic result, where each model uses Tensors with three hidden layers (500x500x500, $l=3$) and the models share the embedding and the first hidden layers ($t=1$).  MTL consistently gives further intrinsic gain compared to tensors. More sharing provides an extra gain for SrcEn as shown in the last column. Note that we only experiment with different $l$ and $t$ for SrcEn and not for HypEn because the models in HypEn have different input sets. In our experiments, further sharing and more hidden layers resulted in no further gain. In total, we see a consistent positive effect in intrinsic evaluation from the tensor networks and multitask learning.

\begin{table*}[htb]
\begin{center}
\begin{tabular}{| l | r | r | r | r | r | r |}
\hline
 \multirow{2}{*}{Feature set} & \multicolumn{3}{c|}{AR-EN} & \multicolumn{3}{c|}{ZH-EN} \\
  \cline{2-7}
  \cline{2-7}
                    & NN & Tensor & MTL & NN & Tensor & MTL \\
\hline
$R_1$: Baseline Features & \emph{43.2} & 43.9 & - & \emph{30.2} & 30.8 & - \\
$R_2$: $R_1$ + HypEn & 44.4 & 44.4 & 44.5 & 30.5 & 31.5 & 31.3 \\
$R_3$: $R_1$ + SrcEn & 44.3 & 44.9 & 45.5 & 30.7 & 31.5 & 31.9 \\
$R_4$: $R_1$ + HypEn + SrcEn & 44.7 & 45.3 & \textbf{45.7} & 30.9 & 31.8 & \textbf{32.0} \\
\hline

\end{tabular}
\end{center}
\caption{Experimental results to investigate the effects of the new features, DTN and MTL. The top part shows the BOLT results, while the bottom part shows the NIST results. The best results for each conditions and each language-pair are in \textbf{bold}. The baselines are in \emph{italics}. \label{t_mt_bolt}.}
\end{table*}
Moving on to MT evaluation, we summarize the experiments showing the impact of Tensors and MTL in Table~\ref{t_mt_bolt}.
For MTL, we use $L=3,t=2$ since it gives the best intrinsic score. Employing tensors instead of regular neural networks gives a significant and consistent positive impact for all models and language pairs. For the system with the baseline features, we use the tensor architecture for both the joint model and the lexical translation model of Devlin et al. resulting in an improvement of around 0.7 BLEU points, and showing the wide applicability of the tensor architecture. On top of this improved baseline, we also observe an improvement of the same scale for other models (column ``Tensor''), except for HypEn features in AR-EN experiment. Moving to MTL experiments, we see improvements, especially from SrcEn features. MTL gives around 0.5 BLEU point improvement for AR-EN and around 0.4 BLEU point for ZH-EN. When we employ both HypEn and SrcEn together, MTL gives around 0.4 BLEU point in AR-EN and 0.2 BLEU point in ZH-EN. In total, our work results in an improvement of 2.5 BLEU point for AR-EN and 1.8 for BLEU point in ZH-EN on top of the best results in \cite{devlin-EtAl:2014:P14-1}.

\subsection{NIST OpenMT12 \label{sec:nist_results}}
Our NIST system is compatible with the OpenMT12 constrained track, which consists of 10M words of high-quality parallel training for Arabic, and 25M words for Chinese. The n-gram LM is trained on 5B words of data from the English GigaWord. For test, we use the ``Arabic-To-English Original Progress Test'' (1378 segments) and ``Chinese-to-English Original Progress Test + OpenMT12 Current Test'' (2190 segments), which consists of a mix of newswire and web data. All test segments have 4 references. Our tuning set contains 5000 segments, and is a mix of the MT02-05 eval set as well as additional held-out parallel data from the training corpora.

We report the experiments for the NIST condition in Table~\ref{t_mt_nist}. In particular, we investigate the impact of deploying our new features (column ``Feat'') and demonstrate the effects of the tensor architecture (column ``Tensor'') and multitask learning (column ``MTL''). As shown the results are inline with the BOLT condition where we observe additive improvements from adding our new features, applying tensor network and multitask learning. On Arabic-English, we see a gain of 2.7 BLEU point and on Chinese-English, we see a 1.9 BLEU point gain. We also report the mixed-cased BLEU scores for comparison with previous best published results, i.e. \newcite{devlin-EtAl:2014:P14-1} report 52.8 BLEU for Arabic-English and 34.7 BLEU for Chinese-English. Thus, our results are around 1.3-1.4 BLEU point better. Note that they use additional rescoring features but we do not.

\begin{table}
\begin{tabular}{| l | l | l | l | l | l |}
\hline
&  Base. & Feat & Tensor & MTL \\
\hline
AR-EN & \emph{53.7} & 55.4 & 55.9 & 56.4  \\
mixed-case & \emph{51.8} & 53.1 & 53.7 & 54.1 \\
\hline
ZH-EN & \emph{36.6} & 37.8 & 38.2 & 38.5 \\
mixed-case &  \emph{34.4} & 35.5 & 35.9 & 36.1 \\
\hline
\end{tabular}
\caption{Experimental results for the NIST condition. Mixed-case scores are also reported. Baselines are in \emph{italics}.\label{t_mt_nist}}
\end{table}

%% file: related_work.tex
\section{Related Work}

Our work is most closely related to \newcite{devlin-EtAl:2014:P14-1}. They use a simple feedforward neural network to model two important MT features: A joint language and translation model, and a lexical translation model. They show very large improvements on Arabic-English and Chinese-English web forum and newswire baselines. We improve on their work in 3 aspects. First, we model more features using neural networks, including two novel ones: a joint model with offset source context and a translation context model. Second, we enhance the neural network architecture by using tensor layers, which allows us to model richer interactions. Lastly, we improve the performance of the individual features by training them using multitask learning. In the remainder of this section, we describe previous work relating to the three aspect of our work, namely MT modeling, neural network architecture  and model learning. 


The features we propose in this paper address the major aspects of SMT modeling that have informed much of the research since the original IBM models \cite{Brown:1993}: lexical translation, reordering, word fertility, and language models. Of particular relevance to our work are approaches that incorporate context-sensitivity into the models \cite{carpuat-wu:2007:EMNLP-CoNLL2007}, formulate reordering as orientation prediction task \cite{tillman:2004:HLTNAACL} and that use neural network language models \cite{bengio2003,schwenk2010,schwenk2012}, and incorporate source-side context into them \cite{devlin-EtAl:2014:P14-1,auli2013,le2012,schwenk2012}.

Approaches to incorporating source context into a neural network model differ mainly in how they represent the source sentence and in how long is the history they keep. In terms of representation of the source sentence, we follow \cite{devlin-EtAl:2014:P14-1} in using a window around the {\em affiliated} source word. To name some other approaches, \newcite{auli2013} uses latent semantic analysis and source sentence embeddings learned from the recurrent neural network; \newcite{sundermeyer-EtAl:2014:EMNLP2014} take the representation from a bidirectional LSTM recurrent neural network; and \newcite{kalchbrenner-blunsom:2013:EMNLP} employ a convolutional sentence model. For target context, recent work has tried to look beyond the classical \emph{n}-gram history. \cite{auli2013,sundermeyer-EtAl:2014:EMNLP2014} consider an unbounded history, at the expense of making their model only applicable for N-best rescoring. Another recent line of research \cite{bahdanauChoBengio,Ilya_NIPS2014_5346} departs more radically from conventional feature-based SMT and implements the MT system as a single neural network. These models use a representation of the whole input sentence. 

We use a feedforward neural network in this work. Besides feedforward and recurrent networks, other network architectures that have been applied to SMT include convolutional networks \cite{kalchbrenner-grefenstette-blunsom:2014:P14-1} and recursive networks \cite{SocherEtAl2011:RNN}. The simplicity of feedforward networks works to our advantage. More specifically, due to the absence of a feedback loop, the feedforward architecture allows us to treat individual decisions independently, which makes parallelization of the training easy and the querying the network at decoding time straightforward. The use of tensors in the hidden layers strengthens the neural network model, allowing us to model more complex feature interactions like collocation, which has been long recognized as important information for many NLP tasks (e.g. word sense disambiguation \cite{Lee:2002}). The tensor formulation we use is similar to that of \cite{conf/interspeech/YuDS12,Hutchinson:2013:TDS:2498740.2498878}. Tensor Neural Networks have a wide application in other field, but have only been recently applied in NLP \cite{NIPS2013_5028,pei-ge-chang:2014:P14-1}. To our knowledge, our work is the first to use tensor networks in SMT.

Our approach to multitask learning is related to work that is often labeled joint training or transfer learning. To name a few of these works, \newcite{finkel-manning:2009:NAACLHLT091} successfully train name entity recognizers and syntactic parsers jointly, and Singh et al. \shortcite{Singh:2013} train models for coreference resolution, named entity recognition and relation extraction jointly. Both efforts are motivated by the minimization of cascading errors. Our work is most closely related to \newcite{Collobert:2008,Collobert:2011}, who apply multitask learning to train neural networks for multiple NLP models: part-of-speech tagging, semantic role labeling, named-entity recognition and language model variations.

%% file: discussion.tex
\section{Conclusion}
This paper argues that a relatively simple feedforward neural network can still provides significant improvement to Statistical Machine Translation (SMT). We support this argument by presenting a multi-pronged approach that addresses modeling, architectural and learning aspects of pre-existing neural network-based SMT features. More concretely, we paper present a new set of neural network-based SMT features to capture important translation phenomena, extend feedforward neural network with tensor layers, and apply multitask learning to integrate the SMT features more tightly. Empirically, all our proposals successfully produce an improvement over state-of-the-art machine translation system for Arabic-to-English and Chinese-to-English and for both BOLT web forum and NIST conditions. 
Building on the success of this paper, we plan to develop other neural-network-based features, and to also relax the limiteation of current rule extraction heuristics by generating translations word-by-word.